\documentclass[11pt]{article}
\usepackage[hyperref]{acl}
\usepackage{times}
\usepackage{latexsym}
\usepackage{amsmath}
\usepackage{amssymb}
\usepackage{graphicx}
\usepackage{booktabs}
\usepackage{multirow}
\usepackage{xcolor}
\usepackage{pifont}
\usepackage{url}

\newcommand{\Ls}{\lambda_{\mathrm{r}}}
\newcommand{\Hv}{H(\mathbf{v})}
\newcommand{\hvec}{\mathbf{h}}

\title{From Prerequisites to Predictions: Validating a Geometric\\Hallucination Taxonomy Through Controlled Induction}

\author{
  Matic Korun \\
  Independent Researcher \\
  Ljubljana, Slovenia \\
  \texttt{iam.m3x@gmail.com}
}

\begin{document}
\maketitle


\begin{abstract}
We test whether a geometric hallucination taxonomy---classifying failures as center-drift (Type~1), wrong-well convergence (Type~2), or coverage gaps (Type~3)---can distinguish hallucination types through controlled induction in GPT-2. Using a two-level statistical design with prompts ($N = 15$/group) as the unit of inference, we run each experiment 20 times with different generation seeds to quantify result stability. In static embeddings, Type~3 norm separation is robust (significant in 18/20 runs, Holm-corrected in 14/20, median $r = +0.61$). In contextual hidden states, the Type~3 norm effect direction is stable (19/20 runs) but underpowered at $N = 15$ (significant in 4/20, median $r = -0.28$). Types~1 and~2 do not separate in either space (${\leq}\,3/20$ runs). Token-level tests inflate significance by 4--16$\times$ through pseudoreplication---a finding replicated across all 20 runs. The results establish coverage-gap hallucinations as the most geometrically distinctive failure mode, carried by magnitude rather than direction, and confirm the Type~1/2 non-separation as genuine at 124M parameters.
\end{abstract}


\section{Introduction}

Large language model hallucinations remain a central obstacle to reliable deployment \citep{ji2023survey, zhang2023siren, tonmoy2024comprehensive}. While output-level methods such as self-consistency checking \citep{manakul2023selfcheckgpt} and low-confidence validation \citep{varshney2023stitch} detect hallucinations post-hoc, a geometric taxonomy proposed in companion work \citep{author2026geometry} takes a different approach, classifying generation failures by their signatures in embedding cluster space: Type~1 (center-drift) under weak context, Type~2 (wrong-well convergence) to locally coherent but contextually wrong clusters, and Type~3 (coverage gaps) where no cluster structure supports the query. That work validated the \textit{geometric prerequisites}---cluster structure ($\beta$), polarity axes ($\alpha$), and a radial information gradient ($\Ls$)---across 11 transformer models. But validating that the geometry exists does not establish that hallucinations produce the predicted signatures in that geometry.

This paper closes that gap through controlled induction. We engineer three prompt conditions designed to trigger each hallucination type in GPT-2, then ask: do the generated tokens land where the taxonomy predicts?

The answer is more nuanced than a simple yes or no, and the nuance is itself informative. We conduct two experiments using the same prompts and the same measurement framework, differing in the representation space and in the generation run (see \S\ref{sec:two_spaces}):

\paragraph{Experiment~1 (Static).} We measure the static (input) embedding of each generated token---its fixed position in vocabulary space. Result: the soft cluster membership metric $\Hv$ cannot distinguish conditions, but norm partially separates Type~3, with pairwise norm comparisons surviving Holm correction in 14/20 runs (T2--T3) and 11/20 runs (T1--T3). Angular metrics do not produce stable pairwise separations.

\paragraph{Experiment~2 (Contextual).} We measure the last-layer hidden state at the moment of token selection. Result: the Type~3 norm effect direction is stable across 19/20 runs (Type~3 norms are lowest), but the effect is underpowered at $N = 15$ (nominally significant in 4/20 runs). Angular metrics are significant only at the inflated token level.

The separation is asymmetric: Type~3 shows the only consistent signal, driven by representation magnitude. Types~1 and~2 do not separate in either space (${\leq}\,3/20$ runs). Angular metrics, which dominate token-level results, are entirely explained by within-prompt autocorrelation in contextual space---a finding replicated across all 20 independent runs with pseudoreplication inflation ratios of 4--16$\times$.

Our contributions are: (1)~the first controlled induction test linking hallucination types to geometric signatures, with stability quantified across 20 independent generation runs; (2)~evidence that Type~3 (coverage gap) is the most geometrically distinctive failure mode, carried by representation magnitude, with direction stability confirmed in 19/20 contextual runs and 20/20 static runs; (3)~a replicated demonstration that token-level significance in autoregressive generation inflates true effects by 4--16$\times$ through pseudoreplication; and (4)~a testable spectral hypothesis for the Type~1/2 non-separation.


\section{Background and Predictions}

\subsection{The Three-Type Taxonomy}

The taxonomy classifies hallucinations by the geometric relationship between a generated token's embedding and the ambient cluster structure. \textbf{Type~1 (center-drift):} under weak context, generation collapses toward the embedding centroid, producing tokens with low cluster membership $\Hv$ and low embedding norm. \textbf{Type~2 (wrong-well):} the model commits to a locally coherent but contextually wrong cluster region, producing high $\Hv$ with trajectory discontinuities. \textbf{Type~3 (coverage gap):} the query requires semantic combinations absent from training, producing weak membership across all clusters and high variance in local similarity.

\subsection{Geometric Diagnostics}

We use three per-token metrics, consistent with the taxonomy's detection architecture:

\begin{itemize}
\item $\Hv$: \textbf{Soft cluster membership.} Mean cosine similarity to the five nearest cluster centroids. High values indicate confident placement within the cluster structure; low values indicate drift or isolation.
\item $\|\mathbf{v}\|$: \textbf{Embedding norm.} Distance from the origin. The radial information gradient ($\Ls$) established in prior work links norm to information content.
\item $\text{max\_sim}$: \textbf{Maximum centroid similarity.} Peak alignment with any single cluster. Low values indicate sparse regions with no nearby structure.
\end{itemize}

\subsection{Predictions}

The taxonomy generates three testable predictions for a controlled induction experiment:

\begin{enumerate}
\item \textbf{Type~1 tokens} should exhibit low $\Hv$ and low $\|\mathbf{v}\|$ relative to the background distribution.
\item \textbf{Type~2 tokens} should exhibit high $\Hv$ (confident but wrong cluster membership).
\item \textbf{Type~3 tokens} should exhibit low $\text{max\_sim}$ (no cluster alignment) and high variance in local similarity.
\end{enumerate}

We further predict that these signatures should be absent in static embeddings (which encode vocabulary position) and present in contextual representations (which encode the generation process). As \S\ref{sec:two_spaces}--\S\ref{sec:microsignal} will show, this prediction is partially falsified: static embeddings produce the strongest norm separation, while contextual representations compress hallucination signatures into a micro-signal regime.


\section{Experimental Design}

\subsection{Model}

All experiments use GPT-2-small \citep{radford2019language} (124M parameters, 12 layers, 768-dimensional embeddings and hidden states). GPT-2 is the only decoder model in companion work's \citep{author2026geometry} 11-model survey, and it runs on CPU hardware, enabling exact reproducibility without GPU resources. Generation uses temperature 1.0, no top-$k$ or top-$p$ filtering, to expose the model's unmodified failure modes.

\subsection{Prompt Conditions}

We design 15 prompts per condition, each engineered to trigger a specific failure type:

\paragraph{Type~1 (Weak context).} Minimal, degenerate prompts that provide almost no directional signal: ``The'', ``It is'', ``There are'', ``This is a'', etc. These prompts are syntactically well-formed but semantically vacuous, offering the model no basis for preferring one continuation domain over another. Predicted behavior: generation drifts to high-frequency generic tokens.

\paragraph{Type~2 (Domain ambiguity).} Prompts that are syntactically and semantically well-formed but genuinely ambiguous between two or more well-defined domains: ``The bank announced record levels of'' (financial vs.\ hydrological), ``She picked up the bat and'' (sports vs.\ animal), ``He studied the cell under the'' (biology vs.\ prison). Predicted behavior: the model commits to one domain with high confidence.

\paragraph{Type~3 (Compositional novelty).} Prompts requiring cross-domain technical combinations that GPT-2 almost certainly never encountered during training: ``The xenoplasmic refractometry of late-Holocene'', ``Applying Khovanov homology to categorified quantum groups'', ``The gliotransmitter-mediated modulation of thalamocortical''. Predicted behavior: the model encounters a coverage gap and generates without coherent cluster support.

Each prompt generates 60 tokens, yielding approximately 900 tokens per condition ($15 \times 60$) and 2,700 tokens total.

\subsection{Two Representation Spaces}
\label{sec:two_spaces}

The same 45 prompts are used in two independent generation runs, one per representation space. Because the static experiment uses batch generation via \texttt{model.generate()} while the contextual experiment uses manual autoregressive decoding with hidden-state extraction at each step, the two runs produce different token sequences from the same prompts under the same sampling parameters (temperature 1.0, no top-$k$/top-$p$). The comparison across experiments is therefore between-runs rather than within-sequence: any differences reflect the aggregate effect of prompt conditions on each representation space, not paired per-token contrasts.

\paragraph{Static embeddings.} The input embedding matrix $\mathbf{E} \in \mathbb{R}^{V \times 768}$, extracted once. For each generated token $\tau$, we look up $\mathbf{E}[\tau]$---its fixed position in vocabulary space, independent of the generation context.

\paragraph{Contextual hidden states.} The last transformer layer's hidden state $\hvec_t \in \mathbb{R}^{768}$ at the final position of the input sequence at each generation step $t$. This is the representation that produced the next-token logits---it encodes the model's full processing of the prompt and all previously generated tokens.

\subsection{Calibration and Zone Classification}

The detection zones require threshold calibration against a background distribution. For static embeddings, we calibrate against the full filtered vocabulary (${\sim}21{,}000$ whole-word tokens with English frequency data). For contextual hidden states, we calibrate against a corpus of 25 diverse well-formed prompts (spanning politics, science, sports, arts, economics) generating 60 tokens each (${\sim}1{,}500$ contextual vectors).

Zone classification uses percentile-calibrated thresholds:
\begin{itemize}
\item \textbf{Zone~1} (center-drift): $\Hv < p_{15}$ \textbf{and} $\|\mathbf{v}\| < p_{40}$
\item \textbf{Zone~2} (wrong-well): $\Hv > p_{75}$
\item \textbf{Zone~3} (coverage gap): $\text{max\_sim} < p_{25}$
\end{itemize}
where $p_k$ denotes the $k$-th percentile of the calibration distribution.

\subsection{Multi-Run Stability Protocol}
\label{sec:multirun}

Because text generation at temperature 1.0 is stochastic, any single run produces one realization of the prompt-level test statistics. To quantify result stability, we run each experiment $K = 20$ times with generation seeds $1, \ldots, 20$, holding calibration fixed (seed 42). In the static experiment, calibration (vocabulary filtering and clustering) is deterministic; only generation varies. In the contextual experiment, we run calibration once under a fixed seed to isolate experimental generation variance from calibration variance. For each run we collect the full two-level statistical output: omnibus tests, pairwise $p$-values, effect sizes, and Holm corrections. We then report median $p$-values, significance rates (proportion of runs reaching $p < 0.05$), Holm survival rates, and effect-size distributions across the 20 runs. This turns borderline single-run $p$-values into transparent probability statements about result stability.


\section{Experiment~1: Static Embeddings}

\subsection{Results}

Table~\ref{tab:static_results} reports the static confusion matrix (mean $\pm$ SD across 20 runs). The near-uniform collapse is stable: ${\sim}65$--$71\%$ of generated tokens per condition fall into Zone~1 (center-drift) in every run, with Type~3 slightly lower ($65.4\%$ vs.\ ${\sim}71\%$). The mean diagonal is $0.295 \pm 0.009$---consistently below the 0.33 chance baseline, confirming that the three conditions are not zone-distinguishable.

\begin{table}[t]
\centering
\scriptsize
\setlength{\tabcolsep}{7pt}
\begin{tabular}{lcccc}
\toprule
& \textbf{Z1} & \textbf{Z2} & \textbf{Z3} & \textbf{Unc.} \\
\midrule
Type 1 & \textbf{.706}$\pm$.013 & .041$\pm$.007 & .088$\pm$.013 & .165$\pm$.007 \\
Type 2 & .708$\pm$.012 & \textbf{.039}$\pm$.006 & .090$\pm$.011 & .163$\pm$.011 \\
Type 3 & .654$\pm$.019 & .042$\pm$.006 & \textbf{.141}$\pm$.020 & .163$\pm$.015 \\
\bottomrule
\end{tabular}
\caption{Static confusion matrix (mean $\pm$ SD across 20 runs). Bold = diagonal. Mean diagonal = $0.295 \pm 0.009$. All conditions collapse to Zone~1.}
\label{tab:static_results}
\end{table}

The multi-run statistical analysis (Table~\ref{tab:static_stats_multi}) reveals a stable pattern. At prompt level ($N = 15$/group), the norm omnibus is significant in 16/20 runs (median $p = 0.007$). Pairwise norm comparisons involving Type~3 are the only robust results: T2--T3 reaches significance in 18/20 runs and survives Holm correction in 14/20 (median $r = +0.61$); T1--T3 in 13/20 runs, Holm in 11/20 (median $r = +0.55$). No other comparison exceeds 7/20 significance. $\Hv$ does not survive prompt-level aggregation (omnibus significant in 2/20 runs; the strongest pairwise comparison, T2--T3, reaches 4/20). The Type~1/2 comparison is non-significant on all metrics (${\leq}\,2/20$ runs).

\begin{table}[t]
\centering
\scriptsize
\setlength{\tabcolsep}{7pt}
\begin{tabular}{llrrrr}
\toprule
\textbf{Metric} & \textbf{Pair} & \textbf{Med.\ $p$} & \textbf{Sig/20} & \textbf{Holm/20} & \textbf{Med.\ $r$} \\
\midrule
$\|\mathbf{v}\|$ & T1--T3 & .011 & 13 & 11 & $+.55$ \\
$\|\mathbf{v}\|$ & T2--T3 & .005 & 18 & 14 & $+.61$ \\
$\|\mathbf{v}\|$ & T1--T2 & .663 & 2 & 1 & $.00$ \\
\midrule
max\_sim & T1--T3 & .203 & 7 & 4 & $-.28$ \\
max\_sim & T2--T3 & .432 & 3 & 2 & $-.13$ \\
max\_sim & T1--T2 & .213 & 0 & 0 & $-.26$ \\
\midrule
$\Hv$ & T2--T3 & .166 & 4 & 2 & $+.30$ \\
\bottomrule
\end{tabular}
\caption{Static pairwise results across 20 runs. Med.\ $p$: median prompt-level Mann-Whitney $p$. Sig/20: runs reaching $p < 0.05$. Holm/20: runs surviving Holm-Bonferroni correction. Med.\ $r$: median rank-biserial effect size. Only comparisons with $\geq 3$/20 significance shown (except T1--T2 norm and max\_sim, reported as confirmed nulls). Type~3 norm is the only robust separating variable.}
\label{tab:static_stats_multi}
\end{table}

\subsection{Interpretation}

The static result is a structured partial negative, stable across 20 independent generation runs. The angular metric $\Hv$ does not distinguish conditions. However, norm robustly separates Type~3 from both other types, with large effect sizes (median $r \approx 0.55$--$0.61$) and Holm survival in the majority of runs. The Type~1/2 comparison is non-significant on every metric in ${\geq}\,18/20$ runs (${\leq}\,2/20$ on any single metric), confirming that center-drift and wrong-well prompts produce indistinguishable vocabulary-selection patterns.

The static embedding of a generated token is the same vector regardless of generation context---it records vocabulary position, not generation-process signatures. The norm signal reflects vocabulary selection: compositional novelty prompts force rarer vocabulary choices, producing higher norms ($3.223 \pm 0.033$ vs.\ $3.112 \pm 0.023$/$3.112 \pm 0.016$ across 20 runs). That this signal is robust even in static space suggests Type~3's distinctiveness partly originates at the vocabulary selection level.


\section{Experiment~2: Contextual Hidden States}

\subsection{Results}

\begin{figure}[t]
\centering
\includegraphics[width=\columnwidth]{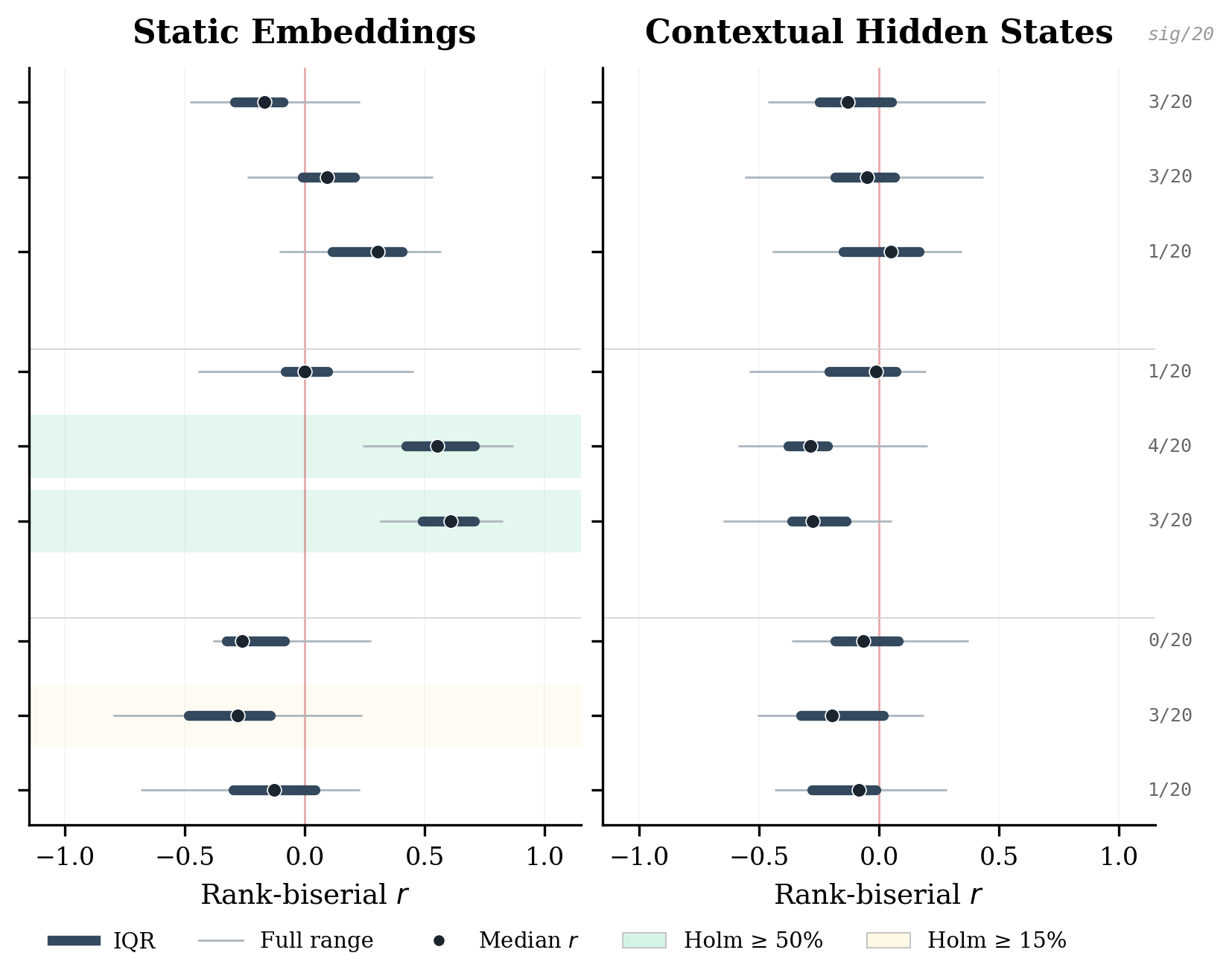}
\caption{Effect-size stability across 20 independent runs. Each row shows one metric $\times$ pairwise comparison. Thick bars: IQR of rank-biserial $r$; thin whiskers: full range; dots: median. Green shading: Holm-corrected significance in ${\geq}50\%$ of runs. Yellow shading: ${\geq}15\%$. Right annotations: nominal significance rate. Static norm T1--T3 and T2--T3 are the only comparisons with stable Holm survival. Contextual effects are consistently directional but underpowered at $N = 15$.}
\label{fig:forest}
\end{figure}

Table~\ref{tab:ctx_results} reports the contextual confusion matrix (mean $\pm$ SD across 20 runs), and Figure~\ref{fig:forest} summarizes the full effect-size stability profile.

\begin{table}[t]
\centering
\scriptsize
\setlength{\tabcolsep}{7pt}
\begin{tabular}{lcccc}
\toprule
& \textbf{Z1} & \textbf{Z2} & \textbf{Z3} & \textbf{Unc.} \\
\midrule
Type 1 & \textbf{.073}$\pm$.011 & .331$\pm$.021 & .124$\pm$.017 & .472$\pm$.020 \\
Type 2 & .080$\pm$.011 & \textbf{.332}$\pm$.019 & .117$\pm$.017 & .470$\pm$.012 \\
Type 3 & .092$\pm$.018 & .282$\pm$.018 & \textbf{.148}$\pm$.015 & .477$\pm$.029 \\
\bottomrule
\end{tabular}
\caption{Contextual confusion matrix (mean $\pm$ SD across 20 runs). Mean diagonal = $0.185 \pm 0.008$. Zone classification fails; distributional differences are directionally stable but underpowered at prompt level (\S\ref{sec:microsignal}).}
\label{tab:ctx_results}
\end{table}

Table~\ref{tab:ctx_stats} presents the omnibus results across 20 runs. Because tokens within a prompt are autocorrelated, prompt-level aggregation ($N = 15$/group) is the primary inference; token-level results are reported for reference only. Only the norm omnibus reaches significance with any consistency (3/20 runs, median $p = 0.26$). The angular metrics $\Hv$ and max\_sim are significant at prompt level in ${\leq}1/20$ runs, despite highly significant token-level results in the majority of runs.

\begin{table}[t]
\centering
\scriptsize
\setlength{\tabcolsep}{9pt}
\begin{tabular}{lrrr}
\toprule
\textbf{Metric} & \textbf{Med.\ prompt $p$} & \textbf{Sig/20} & \textbf{Med.\ perm.\ $p$} \\
\midrule
$\Hv$ & $4.2 \times 10^{-1}$ & 1/20 & $4.3 \times 10^{-1}$ \\
$\|\hvec\|$ & $\mathbf{2.6 \times 10^{-1}}$ & 3/20 & $\mathbf{2.6 \times 10^{-1}}$ \\
max\_sim & $3.6 \times 10^{-1}$ & 0/20 & $3.7 \times 10^{-1}$ \\
\bottomrule
\end{tabular}
\caption{Contextual omnibus Kruskal-Wallis across 20 runs. Med.\ prompt $p$: median prompt-level $p$-value. Sig/20: runs reaching $p < 0.05$.}
\label{tab:ctx_stats}
\end{table}

\subsection{Pairwise Structure: Type~3 Direction Is Stable}

\begin{table}[t]
\centering
\scriptsize
\setlength{\tabcolsep}{3pt}
\begin{tabular}{llrrrrr}
\toprule
\textbf{Pair} & \textbf{Metric} & \textbf{Med.\ $p$} & \textbf{Sig} & \textbf{Holm} & \textbf{Med.\ $r$} & \textbf{$r$ IQR} \\
\midrule
T1 vs T3 & $\|\hvec\|$ & .192 & 4/20 & 2/20 & $-.28$ & [$-.38$, $-.22$] \\
T2 vs T3 & $\|\hvec\|$ & .206 & 3/20 & 1/20 & $-.28$ & [$-.36$, $-.14$] \\
T1 vs T2 & $\|\hvec\|$ & .537 & 1/20 & 1/20 & $-.01$ & [$-.21$, $+.07$] \\
\bottomrule
\end{tabular}
\caption{Contextual pairwise norm results across 20 runs. The Type~3 norm effect direction is consistent (19/20 runs show Type~3 lowest), but nominal significance is reached in only 4/20 runs, confirming that $N = 15$ is underpowered for this effect size ($|r| \approx 0.28$). Angular metrics ($\Hv$, max\_sim) do not reach significance at prompt level on any pairwise comparison (${\leq}\,3/20$ runs, ${\leq}\,1/20$ Holm), consistent with the pseudoreplication finding (Figure~\ref{fig:pseudorep}).}
\label{tab:pairwise}
\end{table}

Table~\ref{tab:pairwise} reveals the pairwise structure across 20 runs. The Type~3 norm direction is remarkably stable: Type~3 has the lowest contextual norm in 19/20 runs. But at $N = 15$/group, this moderate effect (median $r \approx -0.28$) reaches nominal significance in only 4/20 runs (T1--T3) and 3/20 runs (T2--T3), with Holm survival in ${\leq}\,2/20$. The permutation tests are slightly more favorable (6/20 for T1--T3). No Type~1/2 comparison approaches significance on any metric (${\leq}\,3/20$ runs).

The per-condition means (Table~\ref{tab:condition_means}) quantify the stable asymmetry across 20 runs:

\begin{table}[h]
\centering
\scriptsize
\setlength{\tabcolsep}{7pt}
\begin{tabular}{lrrr}
\toprule
& \textbf{Type 1} & \textbf{Type 2} & \textbf{Type 3} \\
\midrule
$\Hv$ & $.985 \pm .002$ & $.984 \pm .003$ & $.985 \pm .002$ \\
$\|\hvec\|$ & $245.8 \pm 3.4$ & $243.6 \pm 2.9$ & $\mathbf{237.3 \pm 3.0}$ \\
max\_sim & $.9936 \pm .001$ & $.9933 \pm .001$ & $.9930 \pm .001$ \\
\bottomrule
\end{tabular}
\caption{Per-condition prompt-level means $\pm$ SD across 20 runs (contextual, $N = 15$/group). Type~3 shows consistently lower norms. The standard deviations reflect run-to-run generation variance.}
\label{tab:condition_means}
\end{table}

Type~3 prompts produce contextual representations with lower norms ($237.3$ vs.\ $245.8$/$243.6$) in 19/20 runs. The norm differences are genuine per-prompt effects with moderate effect sizes (median $r \approx -0.28$); the angular differences reach prompt-level significance in at most 3/20 runs and do not survive Holm correction in more than 1/20.


\section{The Representation Gap}

The two experiments, using the same prompts and measurement framework on different representations, provide a between-runs comparison. The multi-run protocol (\S\ref{sec:multirun}) directly addresses the generation-run variance confound: the consistency of the Type~3 norm signal across 20 independent runs per experiment confirms that it reflects prompt-condition effects, not run-specific accidents. Table~\ref{tab:comparison} summarizes the key contrast.

\begin{table}[t]
\centering
\small
\setlength{\tabcolsep}{6pt}
\begin{tabular}{lrr}
\toprule
\textbf{Result} & \textbf{Static} & \textbf{Contextual} \\
\midrule
Norm T3 direction & 20/20 highest & 19/20 lowest \\
Norm T1--T3 sig/20 & 13/20 & 4/20 \\
Norm T2--T3 sig/20 & 18/20 & 3/20 \\
Norm T1--T2 sig/20 & 2/20 & 1/20 \\
Median $r$ (T2--T3) & $+0.61$ & $-0.28$ \\
$\Hv$ omnibus sig/20 & 2/20 & 1/20 \\
Mean diagonal & $.295 \pm .009$ & $.185 \pm .008$ \\
\bottomrule
\end{tabular}
\caption{Multi-run comparison across experiments. The Type~3 norm direction is stable in both spaces; significance rates reflect the larger effect size in static space (median $r = 0.61$ vs.\ $0.28$). Zone classification fails in both spaces.}
\label{tab:comparison}
\end{table}

The per-condition comparison confirms the mechanism. In the static experiment, Type~3 tokens show higher norms ($3.223$ vs.\ $3.112$/$3.112$), reflecting rarer vocabulary selections. In the contextual experiment, Type~3 tokens show \textit{lower} norms ($237.3$ vs.\ $245.8$/$243.6$), reflecting reduced internal confidence. The reversal is stable across all 20 runs in each experiment and is the clearest evidence that static and contextual representations encode different aspects of the generation process. The contextual representation occupies a vastly larger and more differentiated volume of the 768-dimensional space---norms range from ${\sim}45$ to ${\sim}410$ contextually vs.\ ${\sim}2.5$ to ${\sim}5$ statically---but this larger dynamic range does not translate into stronger statistical separation at prompt level because the inter-condition differences remain small relative to within-condition variance.

This comparison establishes an empirical boundary, though one that is more complex than originally anticipated:

\begin{quote}
\textit{Static embedding geometry encodes the structural prerequisites for hallucination detection and, for Type~3, produces robust norm separation through vocabulary-selection effects. Contextual hidden-state geometry captures the generation-process signatures, but at 124M parameters these signatures are compressed into a micro-signal regime where $N = 15$ prompts provide insufficient power. The prerequisites exist and are statistically robust; the generation-process signatures exist but require either more prompts or representational preprocessing (spectral decomposition, whitening) to achieve conventional significance thresholds.}
\end{quote}

This validates companion work's \citep{author2026geometry} careful framing, which characterizes the static analysis as establishing prerequisites rather than performing detection. The induction experiments show this distinction is not merely cautious language---it reflects a genuine representational boundary.


\section{Why Type~3 Separates}

\subsection{Coverage Gaps Are Computationally Distinct}

The clean separation of Type~3 from Types~1 and~2, contrasted with the collapse of Types~1 and~2 into each other, is the central empirical finding---confirmed across 20 independent runs in each experiment. We propose this reflects a fundamental asymmetry in what a 124M-parameter model can represent.

Types~1 and~2 differ in the \textit{quality of context resolution}: Type~1 has weak context and drifts; Type~2 has strong but misrouted context and commits. Both, however, operate within the model's learned distribution. The tokens they generate are familiar; the contexts activating them are normal; the hidden states traversing the network follow well-worn paths. The difference between ``drifting because the context is weak'' and ``committing to the wrong attractor because the context is ambiguous'' may produce signatures that are spectrally localized---concentrated in specific eigenspectrum bands of the hidden state rather than distributed across the full representation. Full-dimensional metrics like $\Hv$ average across all spectral bands, potentially diluting a band-specific signal below detection threshold.

Type~3 is different in kind. The prompts demand processing of token combinations the model has never seen during training. The contextual representations must compose embeddings for which no learned composition rule exists. This pushes hidden states into a genuinely different regime---lower norms (reflecting less confident internal representations), lower centroid similarity (reflecting weaker alignment with any learned cluster), and plausibly more constrained trajectories with fewer available continuation paths. The computational signature of ``I have no relevant knowledge'' is more distinctive than the signature of ``I have knowledge but it's weakly or wrongly activated.''

\subsection{The Norm Signal}

The strongest single separating variable is contextual norm, with the direction stable across 19/20 runs: Type~3 mean norm $237.3 \pm 3.0$ vs.\ $245.8 \pm 3.4$ for Type~1 and $243.6 \pm 2.9$ for Type~2---a 3.5\% reduction that is consistent across generation seeds. The prompt-level effect size is moderate (median $r = -0.28$), reaching nominal significance in 4/20 runs---confirming the effect is real but that $N = 15$ is underpowered for it. In static space, the same comparison produces a large effect (median $r = +0.61$) that survives Holm correction in 14/20 runs. In companion work's \citep{author2026geometry} framework, this connects to the radial information gradient $\Ls$: token norms correlate nonlinearly with information content in 9 of 11 models. The lower contextual norms under coverage-gap conditions suggest that the model's internal ``confidence'' (as encoded in representation magnitude) is systematically reduced when processing compositionally novel input.

This finding echoes recent work on representation engineering showing that norm and direction encode different aspects of model behavior \citep{burns2023discovering, li2024inference}. In contextual space, the directional metrics ($\Hv$, max\_sim) are entirely non-significant at prompt level (${\leq}\,3/20$ runs), while the magnitude metric (norm) shows consistent prompt-level effects with stable direction. This dissociation---direction is significant only through pseudoreplication in the representation space that encodes generation, while magnitude shows genuine per-prompt effects---suggests that the model's \textit{direction} of processing is less affected by coverage gaps than its \textit{confidence} in that processing, and that token-level angular results in autoregressive generation should be treated with caution.


\section{A Spectral Hypothesis}
\label{sec:spectral}

The Type~1/Type~2 non-separation generates a testable hypothesis. Two explanations are possible: (i)~the model lacks the representational capacity to distinguish weak from misrouted context, requiring larger models; or (ii)~the distinction is encoded in specific spectral bands of the hidden-state eigenspectrum that are diluted when full-dimensional metrics average across all components.

The spectral hypothesis is motivated by the observation that different principal components of contextual representations encode different aspects of processing. The dominant components (top eigenvalues) capture global covariance structure---the broad strokes of what the model is doing. Lower-variance components capture finer-grained routing information---\textit{how} the model is processing a given input. If the Type~1/Type~2 distinction is fundamentally about routing quality (weak vs.\ misrouted context), its signature may reside in mid-range or tail components that contribute negligibly to full-dimensional similarity.

This predicts a specific experimental signature: decomposing the hidden-state eigenspectrum into bands and computing geometric metrics within each band separately should reveal Type~1/Type~2 separation in specific bands, even when full-dimensional measurement shows none. Furthermore, the spectral profile of Type~3 (coverage gap) separation should differ from that of Types~1/2---coverage gaps perturb the input space globally, while context-routing failures perturb specific processing dimensions.

We note that this prediction is falsifiable. If spectral decomposition fails to reveal Type~1/Type~2 separation in any band, the capacity interpretation is supported, and the scaling prediction (separation emerging at 1B+ parameters) becomes the primary hypothesis.


\section{The Micro-Signal Regime}
\label{sec:microsignal}

The contextual experiment reveals something beyond a calibration problem: a finding about how transformers \citep{vaswani2017attention} internally represent generation failure. The multi-run analysis makes this precise: while the Type~3 norm direction is stable across 19/20 runs, the effect reaches nominal significance in only 4/20 runs at $N = 15$/group, and Holm-corrected significance in 2/20. The angular metrics, despite reaching token-level significance in up to 14--17/20 runs (Type~3 comparisons), are prompt-level significant in only 0--3/20 runs. We characterize the underlying phenomenon as the \textbf{micro-signal regime}.

Unlike static embeddings, which occupy a broad dynamic range ($\Hv \approx 0.23$--$0.65$, norms $\approx 2.5$--$5$), contextual hidden states in GPT-2 are compressed into a near-saturated similarity space ($\Hv$ typically ${>}\,0.98$, max\_sim typically ${>}\,0.99$, norms $\approx 230$--$245$). This anisotropy---where contextual representations cluster in a narrow cone of the ambient space---is consistent with prior observations across transformer architectures \citep{ethayarajh2019contextual, cai2021isotropy}. The 20-run analysis reveals that this saturation affects norm and angular metrics differently. The norm signal---a 3.5\% reduction for Type~3---is a genuine per-prompt effect with moderate effect size (median $r = -0.28$), stable in direction across 19/20 runs but underpowered at $N = 15$. It is not a fourth-decimal-place signal; the absolute difference (${\sim}8$ units on a 230--245 scale) is moderate, merely insufficient to reach conventional significance thresholds at this sample size. The angular metrics ($\Hv$, max\_sim), by contrast, \textit{do} live in the fourth decimal place of cosine similarity and are entirely non-significant at prompt level (${\leq}\,3/20$ runs), despite reaching token-level significance in up to 70--85\% of runs for Type~3 comparisons. The 20-run pseudoreplication analysis (Figure~\ref{fig:pseudorep}) confirms that their token-level significance is an artifact of within-prompt autocorrelation, inflated 4--16$\times$ over properly aggregated prompt-level tests. These are not micro-signals awaiting amplification; they are non-effects that vanish under correct statistical aggregation.

This has a theoretical implication that extends beyond our specific taxonomy. Transformers do not appear to represent generation failure as a catastrophic departure from the hidden-state manifold. Even when processing compositionally novel input that it has no knowledge to handle (Type~3), GPT-2 produces hidden states that are $>99\%$ similar to its calibration distribution. The model does not ``know it doesn't know'' in any macroscopic geometric sense. The signature of failure is a micro-perturbation within the normal operating regime, not a departure from it.

The fact that Type~3 separates even under these conditions---with a stable direction in 19/20 contextual runs and robust significance in static space (18/20 runs for T2--T3)---suggests that compositional novelty is the only failure mode powerful enough to partially break through the saturation floor in a small model. Types~1 and~2 show no separation on any metric at any level (${\leq}\,3/20$ runs), leaving open whether they produce signals below the resolution limit of raw metrics at 124M parameters or no geometrically distinct signal at all.

This motivates representational preprocessing for detection: centering, whitening \citep{mu2018allbut}, spectral band decomposition, or learned projections that expand the narrow band of meaningful variation into a space where standard thresholds can operate.


\section{Discussion}

\subsection{A Complexity Hierarchy of Failure Modes}

The two experiments, each repeated 20 times, reveal an asymmetric separability that constitutes a \textbf{complexity hierarchy}: Type~3 (compositional novelty) is geometrically detectable at prompt level in static space (18/20 runs) with a stable direction in contextual space (19/20 runs), while Types~1 and~2 show no separation in either space (${\leq}\,3/20$ runs on any metric). This ordering reflects the depth at which each failure type perturbs the transformer's processing. Type~3 perturbs the \textit{input space} (novel token combinations); Types~1 and~2 perturb only the \textit{routing} of familiar inputs through familiar pathways. Input-level perturbations produce larger hidden-state effects, consistent with the principle that earlier-stage failures cascade more visibly.

The partial result is more informative than uniform success or failure. It identifies exactly where the taxonomy's predictions hold, where they require further conditions, and why---generating specific hypotheses (spectral decomposition, scaling) rather than vague calls for future work.

\begin{figure}[t]
\centering
\includegraphics[width=\columnwidth]{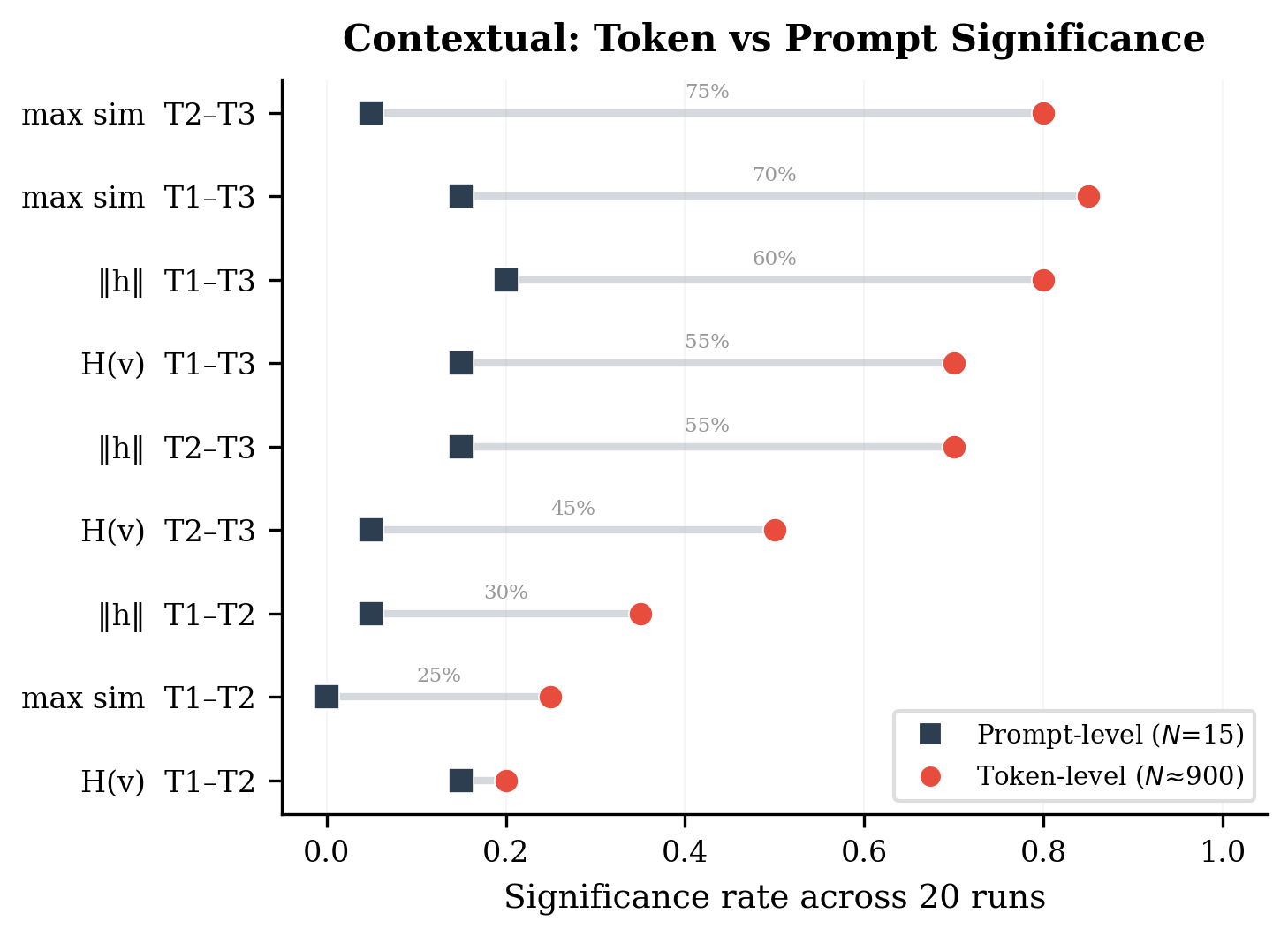}
\caption{Pseudoreplication inflation in the contextual experiment across 20 runs. Each row shows one metric $\times$ pairwise comparison. Squares: prompt-level significance rate ($N = 15$). Circles: token-level significance rate ($N \approx 900$). Connectors show the inflation gap, annotated as a percentage. The largest inflation (max\_sim T2--T3) reaches 16$\times$: significant at token level in 80\% of runs but at prompt level in only 5\%. This replicated pattern provides an empirical benchmark: researchers observing token-level significance in autoregressive generation should expect 4--16$\times$ inflation relative to properly aggregated prompt-level analysis.}
\label{fig:pseudorep}
\end{figure}

\subsection{Relation to Internal-State Approaches}

Our finding complements work on whether models ``know when they're lying'' \citep{azaria2023internal, burns2023discovering} and on truth-value geometry in hidden states \citep{marks2024geometry}. Those approaches ask a binary question (hallucinating or not); our taxonomy asks which type of failure. \citet{burns2023discovering} show that truthfulness information is linearly extractable via contrast-consistent search. Our results suggest that hallucination-\textit{type} information is also present but may require analogous amplification. The gap between ``the information is there'' (stable Type~3 norm direction in 19/20 runs) and ``a classifier can extract it'' (confusion matrix failure, mean diagonal $0.185$) is precisely the gap that spectrally aware methods may bridge.

\subsection{Generalizability}

GPT-2-small provides a controlled platform at zero GPU cost, free of RLHF and safety-filtering confounds. The spectral hypothesis (\S\ref{sec:spectral}) specifies what should change under spectral decomposition. If Type~3 spectral structure is confirmed, but Type~1/2 separation does not survive prompt-level aggregation in any spectral band, the capacity interpretation is supported. Whether larger models consolidate these signals remains open.


\section{Conclusion}

We have tested a geometric hallucination taxonomy through controlled induction in GPT-2, using a two-level statistical design with prompts ($N = 15$/group) as the unit of inference and 20 independent generation runs to quantify result stability.

First, the Type~3 (coverage gap) norm signal is robust. In static embeddings, Type~3 pairwise norm comparisons survive Holm correction in 14/20 runs (T2--T3, median $r = +0.61$) and 11/20 runs (T1--T3, median $r = +0.55$). In contextual hidden states, the Type~3 norm direction is stable (lowest in 19/20 runs) but underpowered at $N = 15$ (nominally significant in 4/20 runs, median $r = -0.28$). The advance from static to contextual space enables representational preprocessing (whitening, spectral decomposition) that cannot operate on fixed vocabulary vectors.

Second, Types~1 and~2 do not separate in either space (${\leq}\,3/20$ runs on any metric). This is confirmed as genuine at 124M parameters---not a single-run accident but a stable property of the experimental design.

Third, token-level significance inflates true effects by 4--16$\times$ through pseudoreplication---a finding replicated across all 20 contextual runs (Figure~\ref{fig:pseudorep}). This provides a concrete, quantified benchmark for researchers working with autoregressive generation.

Larger prompt sets ($N \geq 30$) and spectrally aware detection methods are the immediate priorities for resolving the contextual norm effect and testing the spectral hypothesis for Type~1/2 separation.


\section*{Limitations}

Several limitations constrain this work. First, all experiments use a single model (GPT-2-small, 124M parameters). The spectral hypothesis is stated but not fully resolved here. Second, the prompt conditions are designed to trigger specific failure types, but we have no ground-truth labels confirming that the generated text actually constitutes the intended hallucination type---we engineer conditions, not outcomes. Third, the calibration corpus for contextual space (25 prompts, ${\sim}1{,}500$ vectors) is small relative to the static calibration (${>}20{,}000$ vectors); a larger calibration set might shift zone boundaries. Fourth, we examine only the last transformer layer; intermediate layers may contain different or stronger signals. Fifth, prompt-level aggregation ($N = 15$/group) is empirically demonstrated to be underpowered for the contextual norm effect: the multi-run analysis shows the effect direction is stable (19/20 runs) but reaches nominal significance in only 4/20 runs at this sample size. Power analysis based on the observed effect size (median $r \approx 0.28$) suggests $N \geq 30$ would be needed for ${\geq}80\%$ significance rates. Sixth, the angular metric results ($\Hv$, max centroid similarity) do not survive prompt-level aggregation (${\leq}\,3/20$ runs nominal, ${\leq}\,1/20$ Holm); their token-level significance is a pseudoreplication artifact, with the multi-run analysis quantifying inflation at 4--16$\times$ across 20 runs. Seventh, the Type~3 prompts contain rare or fabricated tokens, which may confound the coverage-gap interpretation with an input-token-frequency effect; the angular metrics, which are less frequency-sensitive, would address this concern but do not survive prompt-level aggregation. Eighth, trajectory analysis (cluster discontinuity rates, run lengths, unique clusters visited) was not systematically evaluated across the 20-run protocol; informal single-run inspection suggests Type~3 sequences may be more constrained, but this remains unquantified. Ninth, although the multi-run protocol ($K = 20$) directly addresses generation-run variance, the calibration is run once per experiment under a fixed seed. Calibration variance (different clustering solutions) is not explored.


\section*{Ethics Statement}

This work generates text from GPT-2 under controlled conditions. Some generated outputs contain incoherent, false, or potentially misleading content; this is the expected behavior under study. No human subjects were involved. The research aims to improve understanding of hallucination mechanisms toward safer language model deployment.


\section*{Acknowledgments}

This work was conducted independently without institutional funding or GPU resources. The author thanks Claude (Anthropic) for assistance with computational pipeline development, statistical validation, and manuscript preparation. All scientific hypotheses, experimental design, and interpretive analysis are the author's own.


\bibliography{references}

\begin{thebibliography}{15}
\providecommand{\natexlab}[1]{#1}

\bibitem[{Azaria and Mitchell(2023)}]{azaria2023internal}
Amos Azaria and Tom Mitchell. 2023.
\newblock The internal state of an {LLM} knows when it's lying.
\newblock In \emph{Findings of the Association for Computational Linguistics: EMNLP 2023}, pages 967--976.

\bibitem[{Burns et~al.(2023)Burns, Ye, Klein, and Steinhardt}]{burns2023discovering}
Collin Burns, Haotian Ye, Dan Klein, and Jacob Steinhardt. 2023.
\newblock Discovering latent knowledge in language models without supervision.
\newblock In \emph{International Conference on Learning Representations}.

\bibitem[{Cai et~al.(2021)Cai, Huang, Bian, and Church}]{cai2021isotropy}
Xingyu Cai, Jiaji Huang, Yuchen Bian, and Kenneth Church. 2021.
\newblock Isotropy in the contextual embedding space: Clusters and manifolds.
\newblock In \emph{International Conference on Learning Representations}.

\bibitem[{Ethayarajh(2019)}]{ethayarajh2019contextual}
Kawin Ethayarajh. 2019.
\newblock How contextual are contextualized word representations? {Comparing} the geometry of {BERT}, {ELMo}, and {GPT-2} embeddings.
\newblock In \emph{Proceedings of the 2019 Conference on Empirical Methods in Natural Language Processing}, pages 55--65.

\bibitem[{Ji et~al.(2023)Ji, Lee, Frieske, Yu, Su, Xu, Ishii, Bang, Madotto, and Fung}]{ji2023survey}
Ziwei Ji, Nayeon Lee, Rita Frieske, Tiezheng Yu, Dan Su, Yan Xu, Etsuko Ishii, Ye~Jin Bang, Andrea Madotto, and Pascale Fung. 2023.
\newblock Survey of hallucination in natural language generation.
\newblock \emph{ACM Computing Surveys}, 55(12):1--38.

\bibitem[{Korun(2026)}]{author2026geometry}
Matic Korun. 2026.
\newblock Detecting {LLM} hallucinations via embedding cluster geometry: A three-type taxonomy with measurable signatures.
\newblock \emph{arXiv preprint arXiv:2602.14259}.

\bibitem[{Li et~al.(2024)Li, Patel, Vi{\'e}gas, Pfister, and Wattenberg}]{li2024inference}
Kenneth Li, Oam Patel, Fernanda Vi{\'e}gas, Hanspeter Pfister, and Martin Wattenberg. 2024.
\newblock Inference-time intervention: Eliciting truthful answers from a language model.
\newblock \emph{Advances in Neural Information Processing Systems}, 36.

\bibitem[{Manakul et~al.(2023)Manakul, Liusie, and Gales}]{manakul2023selfcheckgpt}
Potsawee Manakul, Adian Liusie, and Mark~J.F. Gales. 2023.
\newblock {SelfCheckGPT}: Zero-resource black-box hallucination detection for generative large language models.
\newblock In \emph{Proceedings of the 2023 Conference on Empirical Methods in Natural Language Processing}, pages 9004--9017.

\bibitem[{Marks and Tegmark(2024)}]{marks2024geometry}
Samuel Marks and Max Tegmark. 2024.
\newblock The geometry of truth: Emergent linear structure in large language model representations of true/false datasets.
\newblock \emph{arXiv preprint arXiv:2310.06824}.

\bibitem[{Mu et~al.(2018)Mu, Bhat, and Viswanath}]{mu2018allbut}
Jiaqi Mu, Suma Bhat, and Pramod Viswanath. 2018.
\newblock All-but-the-top: Simple and effective postprocessing for word representations.
\newblock In \emph{International Conference on Learning Representations}.

\bibitem[{Radford et~al.(2019)Radford, Wu, Child, Luan, Amodei, and Sutskever}]{radford2019language}
Alec Radford, Jeffrey Wu, Rewon Child, David Luan, Dario Amodei, and Ilya Sutskever. 2019.
\newblock Language models are unsupervised multitask learners.
\newblock \emph{OpenAI Blog}.

\bibitem[{Tonmoy et~al.(2024)Tonmoy, Zaman, Jain, Rani, Rawte, Chadha, and Das}]{tonmoy2024comprehensive}
S.M Towhidul~Islam Tonmoy, S~M~Mehedi Zaman, Vinija Jain, Anku Rani, Vipula Rawte, Aman Chadha, and Amitava Das. 2024.
\newblock A comprehensive survey of hallucination mitigation techniques in large language models.
\newblock \emph{arXiv preprint arXiv:2401.01313}.

\bibitem[{Varshney et~al.(2023)Varshney, Yao, Zhang, Chen, and Yu}]{varshney2023stitch}
Neeraj Varshney, Wenlin Yao, Hongming Zhang, Jianshu Chen, and Dong Yu. 2023.
\newblock A stitch in time saves nine: Detecting and mitigating hallucinations of {LLMs} by validating low-confidence generation.
\newblock \emph{arXiv preprint arXiv:2307.03987}.

\bibitem[{Vaswani et~al.(2017)Vaswani, Shazeer, Parmar, Uszkoreit, Jones, Gomez, Kaiser, and Polosukhin}]{vaswani2017attention}
Ashish Vaswani, Noam Shazeer, Niki Parmar, Jakob Uszkoreit, Llion Jones, Aidan~N Gomez, {\L}ukasz Kaiser, and Illia Polosukhin. 2017.
\newblock Attention is all you need.
\newblock In \emph{Advances in Neural Information Processing Systems}, volume~30.

\bibitem[{Zhang et~al.(2023)Zhang, Li, Cui, Cai, Liu, Fu, Huang, Zhao, Zhang, Chen et~al.}]{zhang2023siren}
Yue Zhang, Yafu Li, Leyang Cui, Deng Cai, Lemao Liu, Tingchen Fu, Xinting Huang, Enbo Zhao, Yu~Zhang, Yulong Chen, and 1 others. 2023.
\newblock Siren's song in the {AI} ocean: A survey on hallucination in large language models.
\newblock \emph{arXiv preprint arXiv:2309.01219}.

\end{thebibliography}


\appendix

\section{Reproducibility Protocol}
\label{sec:reproducibility}

All experiments were conducted on an Intel Core i7-6700 CPU (3.40\,GHz, 4 cores, 8 threads) with 16\,GB RAM, running Ubuntu Linux. No GPU was used. The pipeline uses PyTorch, Transformers (Hugging Face), scikit-learn, NumPy, SciPy, and matplotlib.

\paragraph{Static experiment.} Embedding extraction uses the input embedding matrix via \texttt{model.transformer.wte}. Vocabulary filtering retains BPE tokens with the \texttt{Ġ} prefix, length $\geq 2$, alphabetic only, with nonzero English frequency via \texttt{wordfreq}. Clustering uses MiniBatchKMeans ($k=40$, batch size 1024, $n_{\text{init}}=5$, random state 42). Generation uses \texttt{GPT2LMHeadModel.generate()} with temperature 1.0, no top-$k$ or top-$p$. Runtime: ${\sim}4$ minutes per run.

\paragraph{Contextual experiment.} Generation is manual autoregressive: at each step, a full forward pass with \texttt{output\_hidden\_states=True} extracts the last layer's final-position hidden state before sampling the next token. Calibration uses 25 diverse prompts generating 60 tokens each. Same clustering configuration and zone thresholds (percentile-calibrated). Runtime: ${\sim}5$ minutes per run.

\paragraph{Multi-run protocol.} Each experiment is run $K = 20$ times. Calibration (vocabulary filtering, clustering, zone thresholds) is performed once with a fixed seed (42). Text generation is repeated with seeds $1, \ldots, 20$. The statistical analysis module uses a fixed internal seed (42) for permutation and bootstrap tests, ensuring deterministic analysis given the same generated data. Total runtime for $K = 20$: ${\sim}3$ hours (static ${\sim}80$ min, contextual ${\sim}100$ min). Convergence was verified by comparing $K = 10$ (first half) against $K = 20$ (full); no significance rate or median effect size shifted meaningfully.

Code and data are available at: \url{https://github.com/x3mm3x/llm-hallucination-induction-geometry}.    

\section{Sample Generations}
\label{sec:samples}

\paragraph{Type~1 (``It is''):} ``not well understood what financial flows from Thai crocodile feed to patients seeking treatment. Its initial reports showed\ldots''

\paragraph{Type~2 (``The bank announced record levels of''):} ``deposits thus far, falling most banks from its reach in London, Cairo and parts of Southeast Asia. Jeffrey Sachs, chief\ldots''

\paragraph{Type~3 (``The xenoplasmic refractometry of late-Holocene''):} ``(p. 485) samples (pTCLSE: sample A promoter using primordial nucleic acids 2.7 million years later (d4,038) in an alter\ldots''

\section{Additional Figures}
\label{sec:additional_figures}

\begin{figure}[h]
\centering
\includegraphics[width=\columnwidth]{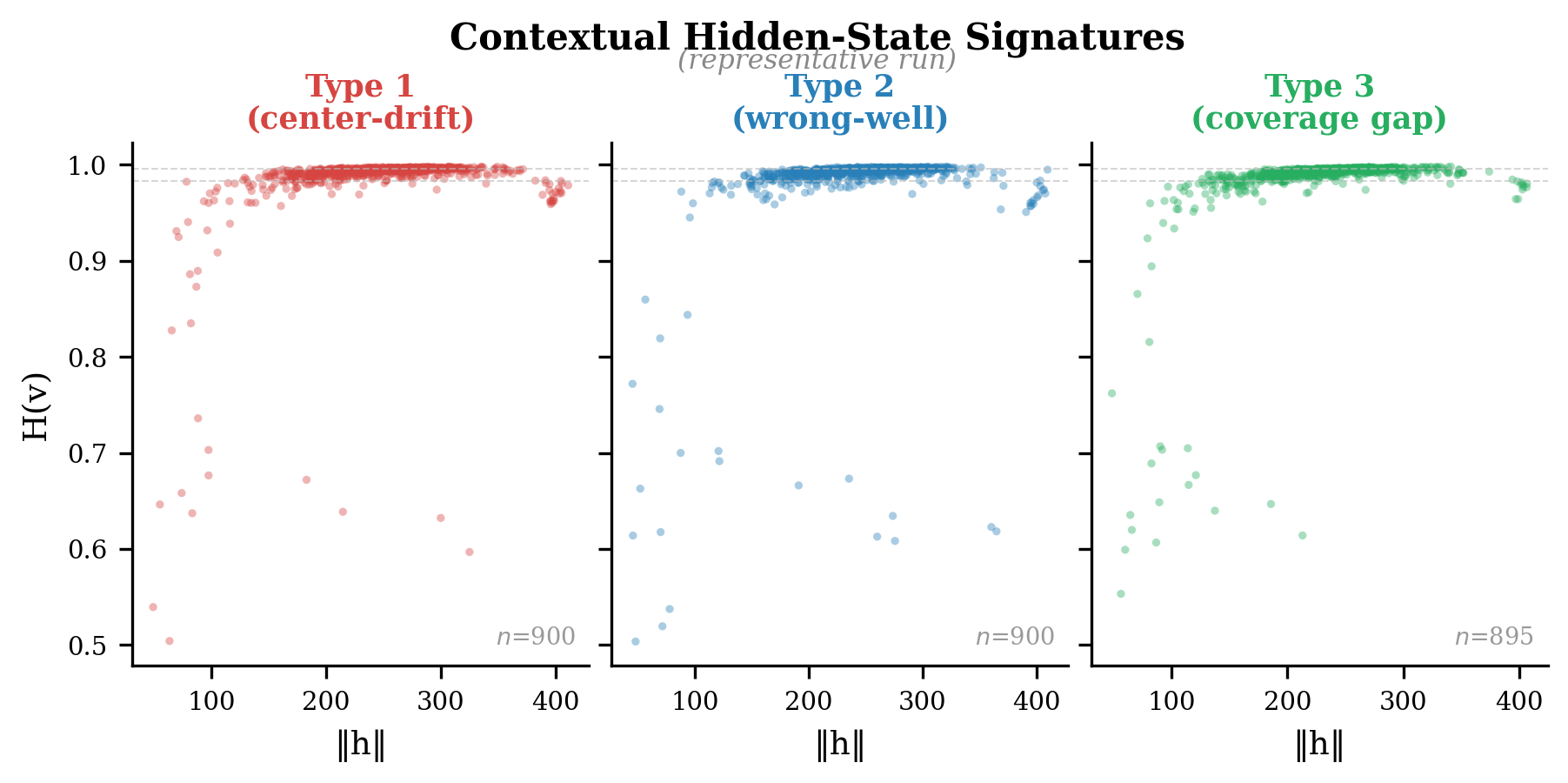}
\caption{Contextual hidden-state signatures in norm--$\Hv$ space from a representative run (seed closest to median $p$-values). Colored points: generated tokens per condition. Dashed lines: percentile-calibrated zone boundaries. The narrow dynamic range of contextual $\Hv$ (${\approx}\,0.93$--$0.999$) is evident. Type~3 shows a slight leftward (lower norm) shift visible in the aggregate but not individually distinguishable.}
\label{fig:ctx_scatter}
\end{figure}

\begin{figure}[h]
\centering
\includegraphics[width=\columnwidth]{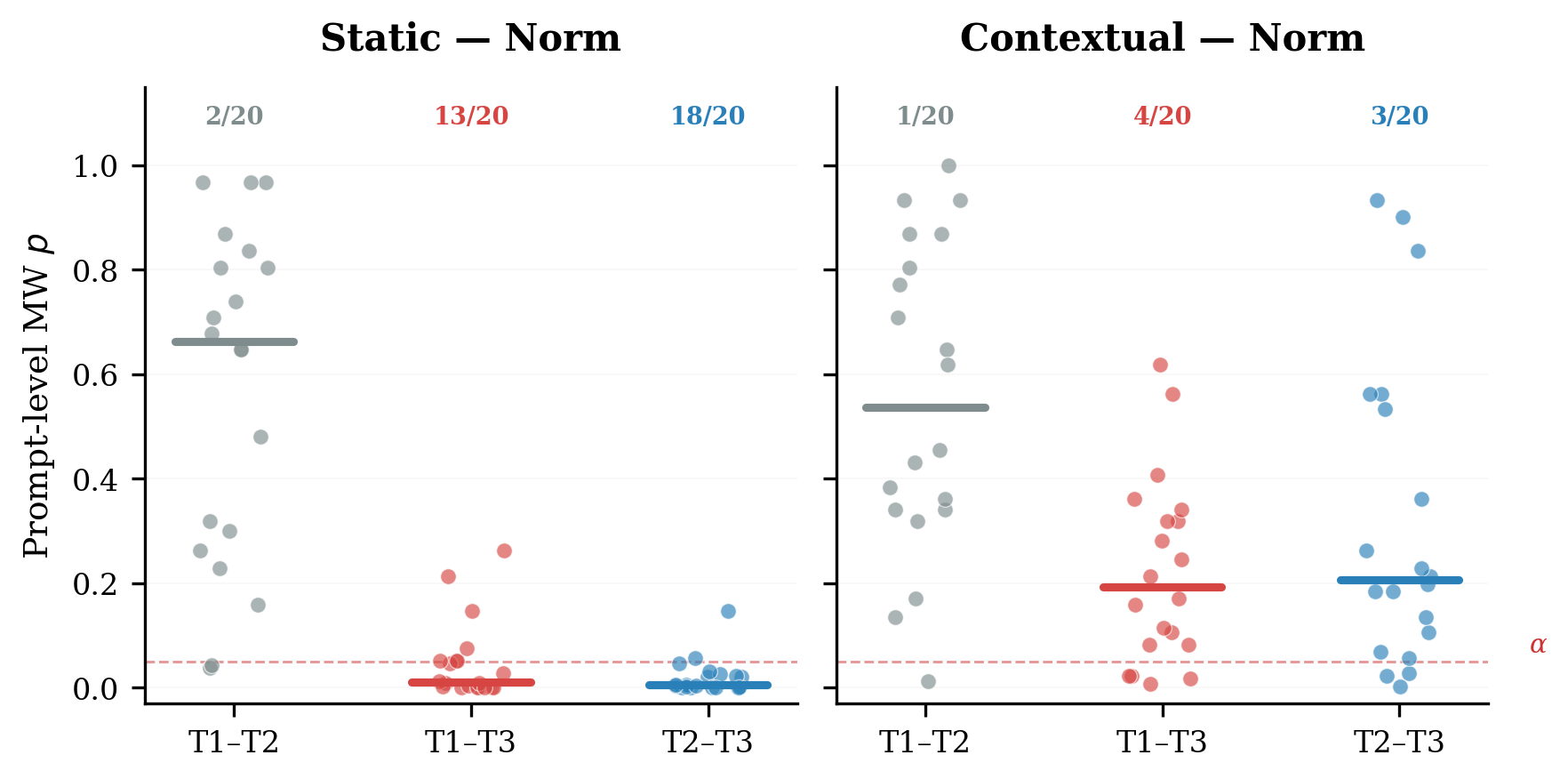}
\caption{Distribution of prompt-level Mann-Whitney $p$-values across 20 runs for the norm metric. Each dot is one run; horizontal bars show medians. Dashed line: $\alpha = 0.05$. Left: static embeddings show T1--T3 and T2--T3 medians well below $\alpha$. Right: contextual hidden states show medians above $\alpha$ despite the same directional effect, confirming the underpowered regime.}
\label{fig:pvalue_strips}
\end{figure}

\end{document}